\documentclass{article}

 \usepackage[preprint,nonatbib]{neurips_2022}

\usepackage[utf8]{inputenc} 
\usepackage[T1]{fontenc}    
\usepackage{hyperref}       
\usepackage{url}            
\usepackage{booktabs}       
\usepackage{amsfonts}       
\usepackage{nicefrac}       
\usepackage{microtype}      
\usepackage{xcolor}         
\usepackage{import}
\usepackage{geometry}
\usepackage{amsmath,amssymb} 
\usepackage{array, caption, floatrow, tabularx, makecell, booktabs}%
\setcellgapes{3pt}
\usepackage{amssymb}
\usepackage{pifont}
\newcommand{\cmark}{\ding{51}}%
\newcommand{\xmark}{\ding{55}}%
\usepackage{multirow}
\usepackage{graphicx}
\usepackage[maxnames=99]{biblatex}
\usepackage{float}
\floatstyle{plaintop}
\restylefloat{table}
\usepackage{wrapfig}
\usepackage{color, colortbl}
\definecolor{Gray}{gray}{0.95}
\usepackage{algorithm}
\usepackage{algpseudocode}

\bibliography{bibliography}
\title{iBoot: Image-bootstrapped Self-Supervised Video Representation Learning}

\author{%
  Fatemeh Saleh \\
  Samsung AI Cambridge\\
  \texttt{f.saleh@samsung.com} \\
   \And
   Fuwen Tan \\
   Samsung AI Cambridge \\
   \texttt{fuwen.tan@samsung.com} \\
   \AND
   Adrian Bulat \\
   Samsung AI Cambridge \\
   \texttt{adrian.bulat@samsung.com} \\
   \And
   Georgios Tzimiropoulos \\
   Samsung AI Cambridge \\
   \texttt{georgios.t@samsung.com} \\
   \And
   Brais Martinez \\
   Samsung AI Cambridge \\
   \texttt{brais.a@samsung.com} \\
}

\begin{document}

\maketitle

\begin{abstract}

Learning visual representations through self-supervision is an extremely challenging task as the network needs to sieve relevant patterns from spurious distractors without the active guidance provided by supervision. This is achieved through heavy data augmentation, large-scale datasets and prohibitive amounts of compute. \textit{Video} self-supervised learning (SSL) suffers from added challenges: video datasets are typically not as large as image datasets, compute is an order of magnitude larger, and the amount of spurious patterns the optimizer has to sieve through is multiplied several fold. Thus, directly learning self-supervised representations from video data might result in sub-optimal performance. To address this, we propose to utilize a strong image-based model, pre-trained with self- or language supervision, in a video representation learning framework, enabling the model to learn strong spatial and temporal information without relying on the video labeled data. To this end, we modify the typical video-based SSL design and objective to encourage the video encoder to \textit{subsume} the semantic content of an image-based model trained on a general domain. The proposed algorithm is shown to learn much more efficiently (i.e. in less epochs and with a smaller batch) and results in a new state-of-the-art performance on standard downstream tasks among single-modality SSL methods.
\end{abstract}

\section{Introduction}
  
Self-supervised learning (SSL) 
is the task of learning representations directly from data 
without the need of manually-defined annotations. 
SSL has very recently gained widespread attention 
due to a series of transformative developments~\cite{simclr_imcl20,simclrv2_neurips20,mocov2} 
that have turned SSL into a key technology 
on a wide range of computer vision applications.
SSL can be used as a pre-training step,
often resulting in improved performance on downstream tasks
compared to that of pre-training with a fully-supervised method. Furthermore, SSL has been shown to be a key component for few-shot applications~\cite{chen2021pareto, gidaris2019boosting, sslfewshot2020}, zero-shot generalization~\cite{wu2020self}, and semi-supervised learning~\cite{zhai2019s4l} among others.

Despite the fast progress in image-based SSL~\cite{caron2021emerging,mocov2, grill2020bootstrap, he2021masked}, 
its success has not yet been fully matched by video-based SSL techniques. For example, in video-based problems, 
downstream performance when fine-tuning from an SSL pre-trained model
trails behind fine-tunning from 
a pre-trained model learned through standard supervision. 
This is likely due to the extra challenges introduced by videos, to wit, much higher optimization costs, higher data dimensionality, and smaller dataset sizes.

Recent video SSL methods typically extend the image SSL ones, both in the contrastive~\cite{cvrl_cvpr21,ma2021contrastive} and non-contrastive (via consistency loss) settings~\cite{brave_iccv21}. 
Typically, these methods either sample different temporal clips within a sequence and extend standard image augmentations to video to create the positive pairs, or extend into multiple modalities so as to integrate optical flow, audio, or text.
For instance, a direct extension of BYOL~\cite{grill2020bootstrap} is the state-of-the-art for single-modality video SSL~\cite{feichtenhofer2021large}, with the only difference being on how to sample different views within a sequence.

In this work, we propose \emph{iBoot}, the ``Image-bootstrapped Video Representation Learning''. We look into an aspect surprisingly overlooked by prior work, that is, image representation learning methods have learnt to represent images expressively and we argue one can skip re-doing this for learning spatial aspects of the video data. Therefore, we propose to learn a representation that rather than replacing it, expands on the frame representation. We achieve this by making sure the video encoder subsumes the frame representation provided by a strong image-based foundation model trained with self-supervision~\cite{caron2021emerging} or natural language supervision such as CLIP~\cite{clip_icml21}. We empirically show that iBoot, which is fundamentally different in spirit compared to existing video SSL methods~\cite{feichtenhofer2021large}, performs better at a usually lower computational requirement. 

More specifically, instead of using the widely adopted design wherein two streams of a network (online and target as in~\cite{feichtenhofer2021large}) with the same architecture are trained, we utilize a strong pre-trained image foundation model to serve as the \emph{target network}, guiding the online network (which can be a regular temporal model such as 3D-ResNet~\cite{feichtenhofer2019slowfast}) to learn a strong and expressive spatial representation throughout training. This happens in parallel to the online network's attempt to further learn the temporal persistency over multiple augmented clips of a sequence.

In summary, our contributions are as follows:\\
\textbf{(1)} We propose a simple yet highly effective approach for video SSL that benefits from strong image foundation models~\cite{caron2021emerging,clip_icml21} in a simple pipeline, outperforming existing video SSL methods on various downstream tasks. We show that our method converges much faster and can use a smaller batch size, 
enabling training with more widely-available and reasonable compute budgets.
\textbf{(2)} We conduct a thorough exploration of key design aspects that lead to a well-performing realization of the proposed idea, including, among others, (a) various architectures (ResNet, ViT), and (b) various types of supervision (fully supervised on ImageNet, self-supervised (SwaV, DINO) and weakly supervised (CLIP)) for training the image-based model. (3) We show that our best trained model outperforms the state-of-the-art method of~\cite{feichtenhofer2021large} on various downstream tasks, e.g., by 1.8\% in linear evaluation and by 2.9\% in semi-supervised classification (1\% labels) on Kinetics-400 dataset.

\section{Related Work}

With the availability of huge amounts of unlabeled data, 
self-supervised representation learning has become an important technique, 
allowing machines to learn useful representation from such source of information. 
This leads to a general feature representation learning which benefits downstream visual recognition tasks. 
Great progress has been made towards visual SSL by focusing on discriminative approaches which treat visual representation learning in a similar fashion to supervised scenario, but where both the samples and labels are derived from an unlabeled data.
A number of these methods rely on handcrafted pretext tasks such as image colorizing~\cite{larsson2016learning, zhang2016colorful} or inpainting~\cite{pathak2016context}
for image representation learning and spatio-temporal ordering~\cite{lee2017unsupervised,misra2016shuffle,wei2018learning,xu2019self}, tracking patches or pixels~\cite{wang2015unsupervised,wang2019learning}, and predicting the playback speed~\cite{benaim2020speednet} 
for video representation learning.
Although effective to some extent, these approaches mostly rely on the heuristic design of auxiliary tasks which limit the generality of the learned representations.

An alternative to defining pretext tasks is to learn discriminative features based on contrastive learning in the latent space~\cite{simclr_imcl20,simclrv2_neurips20,mocov2,HardNegMine_neurips20}. To this end, a contranstive loss~\cite{van2018representation} is employed to learn how to pull together the positive pairs, coming from different views of the same image, and push apart the negative pairs, i.e., from different images.
For example, SimCLR~\cite{simclr_imcl20} constructs positives through different augmentations (views) of the same data example and negatives from other samples within a batch, and contrasts them in
the latent space. 
Instead, MoCo~\cite{mocov2} maintains a queue of negative samples and turns one branch into a momentum encoder to improve the consistency of the queue.
Contrastive learning was shown to also be successful for video SSL~\cite{behrmann2021long,han2020memory,hu2021contrast,huang2021self,jenni2021time,li2021motion,lin2021self,cvrl_cvpr21,wang2021removing}. 

Furthremore, compared to images, videos have
other sources of information, such as audio and motion, which can provide contrastive approaches with a variety of modalities. In particular, a contrastive loss has been widely utilized for cross-modal discrimination of video from audio and/or optical flow~\cite{akbari2021vatt,alwassel2020self,arandjelovic2017look,korbar2018cooperative,ma2021contrastive,morgado2021audio,patrick2020multi,piergiovanni2020evolving,patrick2021compositions,patrick2021space}. Adding other modalities besides a visual one has been of interest in recent years in the image domain as well. As an example,~\cite{clip_icml21} utilizes natural language supervision for image representation learning.

Recent works have shown that visual representations can be learned without discriminating between samples and instead focus only on learning from views of the same sample~\cite{swav2020,caron2021emerging,simsiam_cvpr21,grill2020bootstrap}. In particular, BYOL~\cite{grill2020bootstrap} and DINO~\cite{caron2021emerging} directly predict the output of one view from another view in a Siamese network in which one branch is a momentum encoder. Instead, SwAV~\cite{swav2020} predicts cluster assignments of the representation of the opposite view, rather than the representation itself.
Consistency loss has been also employed in video representation learning as in~\cite{feichtenhofer2021large,brave_iccv21}. 
In these approaches, in addition to image augmentations, different temporal clips within a sequence are sampled and used to build positive pairs, forcing a consistent representation across different video segments. 

Recently, inspired by BERT~\cite{devlin2018bert}, which was developed for masked language modeling, masked image modeling has been
proposed for visual representation learning~\cite{beit_iclr22,he2021masked,zhou2021ibot}. To this end, during pre-training, they randomly mask some proportion of image patches and feed the corrupted
input to a Transformer model. The model learns to recover the visual tokens~\cite{beit_iclr22, zhou2021ibot} or pixels~\cite{he2021masked}. 
The pre-training task of masking and prediction has been recently used for
video SSL as well, following a similar strategy of masking video blocks and aim to reconstruct either a HOG representation of the masked blocks~\cite{wei2021masked}, or in~\cite{vimpac_arxiv21}, discretized video tokens generated via an offline VQ-VAE~\cite{van2017neural}.

Unlike the aforementioned methods, we take an alternative approach to efficiently learning video representations. In this work, we build our approach upon the motivation of leveraging strong pre-trained image SSL foundation models to help the video SSL model to better learn the spatial aspect of the videos. We note that
learning video representations by means of leveraging image representations has been previously investigated~\cite{girdhar2019distinit}. In particular,~\cite{girdhar2019distinit} proposes to use one or several image models trained through supervision (e.g. ImageNet-1k, PlaceNet) to assist in learning a spatial-temporal representation. To this end, the image model is used to obtain a prediction for each of the frames, and the soft predictions are used to supervise the video encoder. However, unlike~\cite{girdhar2019distinit}, we frame our method purely in the self-supervised learning realm, do so within a state-of-the-art modern SSL method, and use a methodology that induces temporally-aware representations rather than representations that only work well for spatially datasets.

\section{Proposed Method}

\begin{figure*}[!t]
  \centering
		\includegraphics[width=1\linewidth]{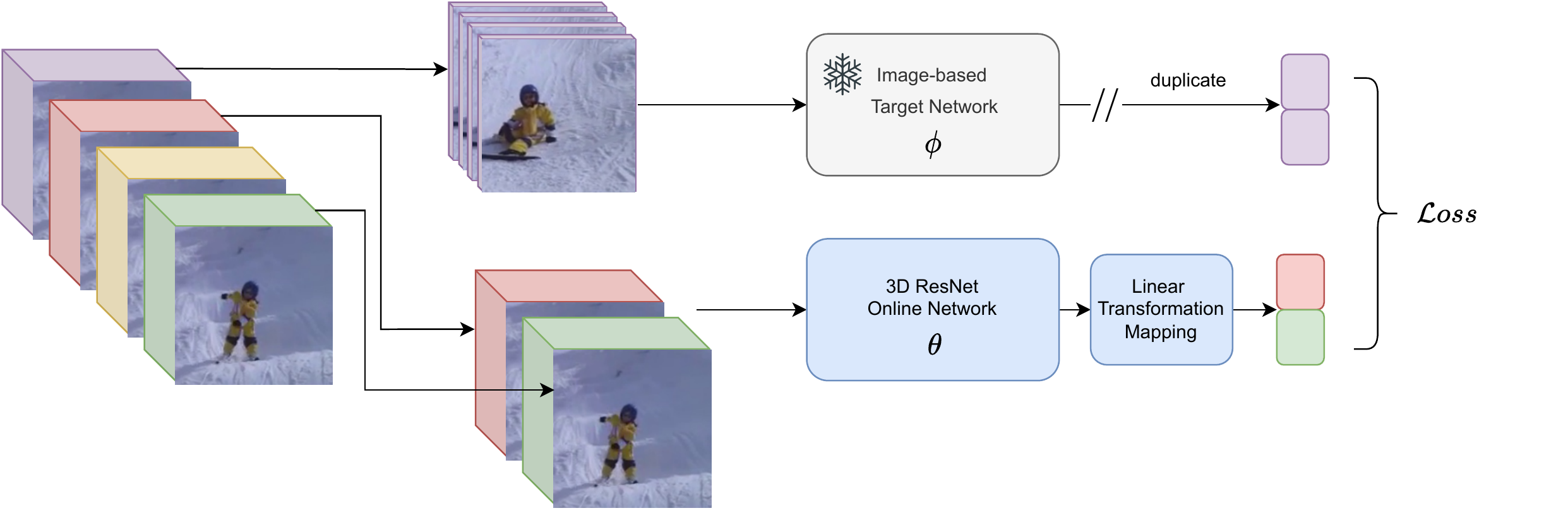}
\caption{An overview of iBoot: during training, iBoot encourages the features of $k$ clips computed by online network to be similar to the reference features computed by the frozen target network.}
\label{fig:pipeline}
\vspace{-5pt}
\end{figure*}

\label{sec:method}

We develop our approach upon $\rho$BYOL (the best performing method from the ones described in~\cite{feichtenhofer2021large}) by extending it to work in synergy with a pre-trained image-based model, rather than in a purely video self-supervised manner (as proposed in~\cite{feichtenhofer2021large}). $\rho$BYOL utilizes two neural networks with the same architecture, the online and target networks, to learn the visual representation. The online network is defined by a set
of weights $\theta$ and the target network's parameters $\theta_m$ are the exponential moving average of the online parameters $\theta$. BYOL minimizes negative cosine similarity between the online and target representations, each taking as input two (spatially and temporally) augmented views (also refer to as clips) of the same video. In this work, the proposed iBoot extends $\rho$BYOL by capitalizing on the benefits of strong image-based foundation models.

\begin{figure}[t]
\setlength{\tabcolsep}{1pt}
\centering
\small
\scalebox{0.95}{
\begin{tabular}{c c c c c c c}
cheer leading & flying kite & pole vault & riding elephant & skiing & throwing ball & tying bow tie \\
\includegraphics[width=.14\textwidth]{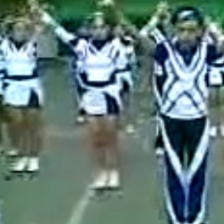} & \includegraphics[width=.14\textwidth]{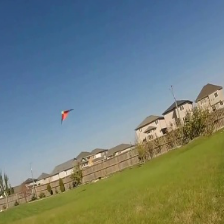} &
\includegraphics[width=.14\textwidth]{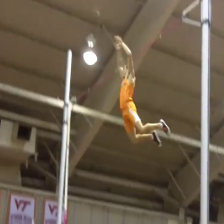} & 
\includegraphics[width=.14\textwidth]{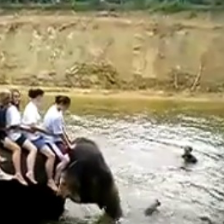} &
\includegraphics[width=.14\textwidth]{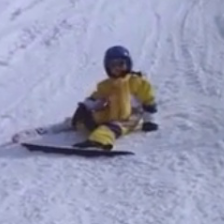} &
\includegraphics[width=.14\textwidth]{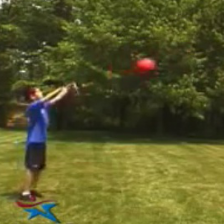} &
\includegraphics[width=.14\textwidth]{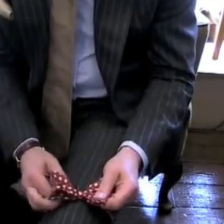} \\
 \includegraphics[width=.14\textwidth]{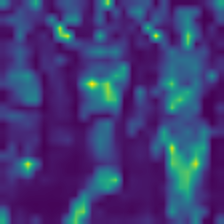} &
 \includegraphics[width=.14\textwidth]{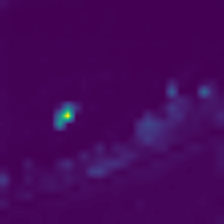} &
 \includegraphics[width=.14\textwidth]{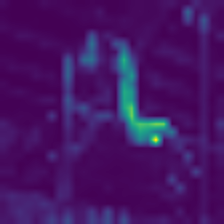} &
 \includegraphics[width=.14\textwidth]{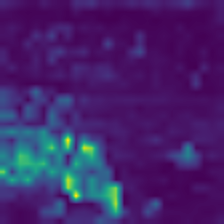} &
 \includegraphics[width=.14\textwidth]{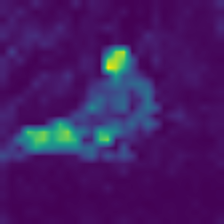} &
 \includegraphics[width=.14\textwidth]{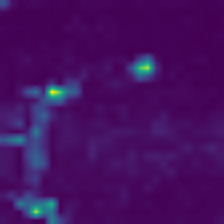} &
 \includegraphics[width=.14\textwidth]{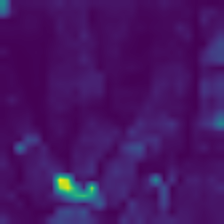}
\end{tabular}}
\vspace{-5pt}
\caption{Attention maps of an ImageNet-1k pre-trained DINO model on Kinetics-400 frames. 
Although the model has not been trained or fine-tuned on any action/video dataset, it correctly attends to the relevant spatial information from the video.
} 
\label{fig:samples}
\end{figure}

More specifically, as illustrated in Figure~\ref{fig:pipeline}, let an image-based pre-trained recognition model, parameterized with $\phi$ to denote our frozen target network. The online network, parameterized with $\theta$, can be a standard video encoder such as 3D ResNet~\cite{feichtenhofer2019slowfast}. To train our model, we utilize different views, $v_{ref}$, $v_{1}$, and  $v_{2}$, randomly sampled and augmented from a sequence. While $v_{ref}$ acts as the input to the target network, we pass $v_{1}$ and $v_{2}$ to the online network. To obtain the reference features of the target network, we simply use the representations (framed-based or average pooled) produced by the pre-trained image-based model, for example the penultimate layer feature from ResNet or the output \texttt{[CLS]} token of the ViT. This way the image-based model can be utilized to provide a rich representation of the frames of the clip. Then, during training, we simply encourage the features of $v_{1}$ and $v_{2}$ produced by the online network to match that of the target one by minimizing the cosine distance loss,
\vspace{-10pt}
\begin{align}
\label{eq:loss}
    \mathcal{L} = - \sum_{k \in \{k^+\}} \text{sim}(q, k) = - \sum_{k\in \{k^+\}} 2 - 2.\frac{\langle q \; , \; k \rangle }{\| q \|_2. \|k\|_2}
\end{align}
where $q = f_{\phi}(v_{ref})$ is the features computed by the frozen target network and $k^+ \in W(f_{\theta}(v_i))$ for $i\in\{1, 2\}$ are the transformed features computed by the online network, wherein $W$ is a linear transformation aiming to map the output dimension of the online network to that of the target one.

Given the above video representation learning setting, we explore the following directions for the image-based model: (a) \textit{Different architectures:} we explore ResNet vs ViT models, and (b) \textit{Different types of supervision:} we explore fully-supervised, self-supervised (i.e. SwAV~\cite{swav2020}, DINO\cite{caron2021emerging}) and weakly-supervised (i.e. CLIP~\cite{clip_icml21}) models.  

One potential caveat is the relative domain gap between the image data it has been trained on and the video data it is applied to. As a sanity check, we empirically show that self-supervised models like DINO provide a general representation capable of capturing the semantic content of the frames in a video. Figure~\ref{fig:samples} illustrates the attention maps of a DINO-ViT (pre-trained on ImageNet-1k) on Kinetics-400 frames. As shown, these attentions maps correspond to coherent semantic regions of the actions of interest while the model has not been trained on any action dataset. This encourages us to perform the extensive exploration described in Sec. \ref{subsec:ablation}.

\textbf{Remark.}
Although there have been great effort in avoiding the collapse phenomenon when pre-training SSL models in a non-contrastive setting, either through incorporating stop gradients~\cite{simsiam_cvpr21}, whitening~\cite{ermolov2021whitening}, or other techniques~\cite{zbontar2021barlow}, our approach suggests a simple yet effective way to handle the collapse by freezing the pre-trained target network. This alleviates the need for further tricks to avoid collapse since iBoot does not update the target network's parameters.

\section{Experiments and Results}
   In this section, we describe the datasets, evaluation settings, and implementation details for pre-training and downstream tasks. We then present the results of an extensive set of ablation studies and finally compare our method to the state-of-the-art video SSL approaches. 

\vspace{-5pt}
\subsection{Pre-training}
\label{sec:imp}
\textbf{Datasets.}
Our self-supervised pre-training is done on Kinetics-400~\cite{kay2017kinetics}(K400) which consists of  $\sim$240k training and $\sim$20k validation video clips in 400 human action categories.
In order to see the effect of using a larger-scale dataset during pre-training, we utilize the Kinetics-600~\cite{carreira2018short}(K600) dataset which has $\sim$392k training videos and $\sim$30k validation videos in 600 classes. It is worth mentioning that we do not use any of these labels during the self-supervised pre-training.

\textbf{Augmentation.}
For data augmentation during the self-supervised pre-training, we follow almost the same setting as in~\cite{feichtenhofer2021large}. 
Given a full-length video, we randomly sample
a clip ($T\times\tau$ frames), thus
the input to the video encoder are $T$ frames subsampled
from the raw clip with a stride of $\tau$ where $T=\tau=8$. Then, for each clip, we apply a consistent spatial augmentation on its frames as well. To this end, spatial cropping is done by randomly resizing the shorter spatial side of a video between [256,320] pixels and takes a random $224\times224$ crop extended over temporal dimension to extract a clip. To each clip, we apply a random horizontal flip,
color distortion, and Gaussian blur following the recent self-supervised approaches~\cite{feichtenhofer2021large,cvrl_cvpr21}. We use the color jittering with the probability of 0.8 and
the color parameters $\left\{ 
\text{brightness, contrast, saturation,
hue}\right\} = \left\{0.2, 0.2, 0.2, 0.05\right\}$.
We also apply random grayscaling with the probability of 0.2. For Gaussian blur we use a spatial kernel with standard-deviation
$\in [0.1, 2.0]$  applied with a probability of 0.5.

\textbf{Implementation details \footnote{iBoot is implemented upon PySlowFast~\cite{fan2020pyslowfast}.}.}
By default, we use a 3DResNet-50, $8\times8$ Slow
pathway~\cite{feichtenhofer2019slowfast} as backbone. The input to the network are clips of $T=8$ frames sampled with stride $\tau=8$ from 64 raw-frames of video. We train the models with the LARS optimizer~\cite{you2017large} for 100 epochs and a batch size of 104 ($13\times8$) distributed
over 8 V100 GPUs for the ablation studies. For the comparison to the state-of-the-art, we increase the number of training epochs to 200 and we use a batch size of 416 ($13\times32$) distributed
over 32 GPUs. Note that synchronized batch normalization across all GPUs is used during training. The learning rate is linearly ramped up during the first 8 epochs to its base value which is $lr = 2.4 \times \frac{bs}{256}$ using linear scaling rule~\cite{goyal2017accurate} and then we decay
the learning rate with a cosine scheduler~\cite{loshchilov2016sgdr}. The weight decay has been set to $10^{-6}$ during training. For the image-based models, we use the publicly released models such as DINO~\cite{caron2021emerging} and CLIP~\cite{clip_icml21}. In particular, for the comparison to the state-of-the-art, we  use the publicly released CLIP ViT-B/16 model~\footnote{Available at \url{https://github.com/openai/CLIP}.}. This model has been pre-trained with the WIT dataset~\cite{clip_icml21} containing about 400M image-text pairs.

\subsection{Evaluation}
\label{subsec:eval}
In this section, we present the details of our evaluation protocols to evaluate our self-supervised video representation learning pipeline. 
Please note that, unless otherwise stated, for all the evaluation settings, we use clips of $T=8$ frames sampled with stride $\tau=8$ from 64 raw-frames of video.

\textbf{Linear evaluation.}
Following existing methods~\cite{feichtenhofer2021large,corp_iccv21,cvrl_cvpr21,brave_iccv21}, we evaluate our approach by training a linear classifier
on frozen features of K400 training set and report the top-1 classification accuracy on K400 validation set. In order to train the linear classifier, we use SGD as our optimizer with momentum set to 0.9. The learning rate starts from 0 and linearly increased during the first 8 epochs to its base
value of 1.0 and then decay it with a cosine scheduler~\cite{loshchilov2016sgdr}. The model is trained with a mini-batch size of 64 for 100 epochs. During testing, we densely sample 10 clips from
each video and apply a 3-crop evaluation following~\cite{feichtenhofer2021large,corp_iccv21, cvrl_cvpr21, brave_iccv21}. 

\textbf{Action classification.}
We report the fine-tuning accuracy on UCF101~\cite{soomro2012ucf101}, HMDB51~\cite{Kuehne11}, and Something-Something V2~\cite{goyal2017something}(SSv2) datasets. For UCF101 and HMDB51, we report results for the ablation studies using
split1 for train/test split. We also report the averaged results over 3 splits of train/test for comparison to previous work, as in Table~\ref{tab:sota}. 
For training the model on UCF101 and HMDB51, we use SGD with the momentum of 0.9 and an initial learning
rate of 0.0025. The models are trained with a mini-batch size of 128 for 196 epochs. During testing, we densely sample 10 clips from
each video and apply a 3-crop evaluation following~\cite{feichtenhofer2021large,corp_iccv21, cvrl_cvpr21, brave_iccv21}. 
SSv2 dataset contains
about 168k training videos and 24k validation videos in
174 video categories. What makes this dataset different from UCF101 or HMDB51 is the fact that many action categories in this dataset
shares very similar background and object appearance, thus learning temporal information is crucial for the model to classify different actions.
For training, we use SGD with the momentum of 0.9 and weight decay of $10^{-4}$. The learning rate starts from 0.08 and linearly ramped up during the first 3 epochs to its base
value of 0.15. Then we decay the learning rate with a cosine scheduler~\cite{loshchilov2016sgdr}. 
The models are trained with a mini-batch size of 64 for 40 epochs. During testing, unless otherwise stated, we follow~\cite{feichtenhofer2021large} and sample 1 clip from
each video and apply a 1-crop evaluation.

\textbf{$k$-Nearest Neighbor evaluation.}
Since both training a linear classifier on frozen features or fine-tuning the features are sensitive to hyperparameters, inspired by~\cite{caron2021emerging, wu2018unsupervised}, we also evaluate the quality of features with a  weighted nearest neighbor classifier ($k$-NN) and report top-1 classification accuracy on the validation set of K400. In particular, we use the frozen features of
the pre-trained model computed from the training set of K400. The nearest neighbor
classifier then matches the feature of a clip from the validation set to the $k$
nearest stored features that vote for the label. Following~\cite{caron2021emerging} we use $k=20$. This is a relatively fast and fair evaluation protocol as it does not rely on any  hyperparameter tuning and the evaluation can be done with a single forward pass over the downstream dataset.

\textbf{Semi-supervised learning.}
To further evaluate the video representations and compare it with the state-of-the-art approaches, we conduct semi-supervised
learning experiments, i.e., fine-tuning the pre-trained network on
small subsets of K400. Following~\cite{cvrl_cvpr21}, we sample 1\% and 10\% videos
from each class in the training set. The settings are similar to the linear evaluation, except
that the initial learning rate is 0.01. We evaluate this experiment on the original K400 validation set.

\subsection{Ablation Studies}
\label{subsec:ablation}
For the ablation studies, for UCF101 and HMDB51, we report top-1 classification accuracy using
split1 for train/test split. We also report the top-1 classification accuracy of a $k$-NN classifier on the K400 dataset for different settings. Please note that unless otherwise stated, all the video SSL training in the ablation studies have been done on a single machine with 8 V100 GPUs for 100 epochs.

\textbf{Effect of image-based pre-training with different level of supervision.}
As discussed in Section~\ref{sec:method}, iBoot utilizes the strong image-based pre-trained models to better learn semantic representations of the video frames. In Table~\ref{tab:image-based}, we compare a model that
has been trained with two branches of video encoder, $\rho$BYOL (the spatio-temporal extension of~\cite{grill2020bootstrap}), with those that utilize the pre-trained image-based models as the target network. In $\rho$BYOL, as in~\cite{feichtenhofer2021large}, the target network's parameters $\theta_m$ are the
exponential moving average of the online parameters $\theta$. As clearly shown in Table~\ref{tab:image-based}, utilizing an image-based pre-trained model is effective for video representation learning, leading to superior performance in all downstream tasks. 
In the last row of the table, we also show the results obtained by utilizing a fully-supervised pre-trained network (DeiT-ViT~\cite{deit_icml21}\footnote{Available at\url{https://github.com/facebookresearch/deit}}) as the target network which interestingly has lower performance compared to its self-supervised counterpart, DINO, and the one using language supervision, CLIP. Similar observation has been made in~\cite{wei2021masked} as well for masked feature prediction. We argue that the degradation in the performance of a model trained with full supervision is due to the fact that class-level supervision targets a much narrower set of semantic concepts than purely self-supervised models such as DINO or vision-language models such as CLIP.

\textbf{Effect of image-based target network architecture.}
According to Table~\ref{tab:image-based}, comparing the results obtained by using SwAV~\cite{swav2020} with Resnet50-w2 and DINO~\cite{caron2021emerging} and CLIP~\cite{clip_icml21} with ViT clearly shows the benefit of using ViT over standard convolutional models even with the same number of parameters. This superiority may come from better feature representation thanks to ViT's self-attention mechanism.

\textbf{Effect of ensembling image-based models.}
The results shown in Table~\ref{tab:multi-image} indicate the effect of using multiple image-based models instead of one \footnote{To train such model, Eq.~\ref{eq:loss} is extended to compute the distance between $k$ with $q$ of each target network.}. While using DINO and CLIP together leads to an improvement in downstream tasks compared to a DINO only setting, it not only offers no improvement over CLIP only setting, but also marginally decreases the accuracy on downstream tasks. This may be due to the fact that CLIP's features are expressive enough while not being in complete agreement with DINO's features. Based on this experiment and the results in Table~\ref{tab:image-based}, we found CLIP to be the best performing image-based target network and thus we use this model in our comparison with the state-of-the-art. It is clear that with the existence of strong image-based models such as CLIP, adding more constraints by involving more image-based models has negative effect on the performance of downstream tasks.

\begin{table}[t]
\renewcommand{\arraystretch}{1.0}
 \small
  \floatsetup{floatrowsep=qquad, captionskip=4pt}
  \begin{floatrow}[2]
    \ttabbox%
    {\begin{tabular}{l c c c c }
    \toprule
             method & Image model &$k$-NN & HMDB & UCF\\
      \midrule
       $\rho$BYOL  & - & 33.2 & 57.7 & 87.3\\ 
       iBoot & SwAV-R50-w2 & 38.8 & 57.4 & 89.0\\ 
       iBoot & DINO-ViT & 48.0 & 59.6 & 89.6 \\ 
       iBoot & CLIP-ViT & 53.1 & 66.5 & 92.6\\ 
       iBoot & DeiT-ViT* & 45.6 & 60.6 &  88.8 \\ 
       \bottomrule
      \end{tabular}}
    {\caption {Effect of different image-based models as target network. (*) denotes the fully-supervised model.}
      \label{tab:image-based}}
    \hfill%
    
    \ttabbox%
    {\begin{tabular}{l c c c }
    \toprule
              Image model & $k$-NN &HMDB & UCF\\
      \midrule
       DINO-ViT & 48.0 & 59.6 & 89.6\\ 
       CLIP-ViT & 53.1 & 66.5 & 92.6 \\ 
       DINO+CLIP & 52.6 & 64.3  & 92.4\\ 
       \bottomrule
      \end{tabular}}
    {\caption {Effect of ensembling image-based models.}
      \label{tab:multi-image}}
  \end{floatrow}
\vspace{-5pt}
\end{table}

\textbf{Effect of additional SSL loss.}
In this experiment, we extend the iBoot pipeline by adding another video encoder branch as an additional target network, parameterized  with $\theta_m$ which is the exponential moving average of the online parameters $\theta$. To train this pipeline, we optimize the parameters via a cosine distance loss for each online-target network pair. According to the results in Table~\ref{tab:ssl-loss}, although adding this SSL loss is slightly effective when using DINO model as the image-based model, it marginally harms the performance when utilizing CLIP as the image-based target network. This experiment clearly shows that using an image-based pre-trained model allows us to skip adding momentum encoder(s), which itself comes at the cost of additional computational overhead. Note that $\rho$BYOL works by enforcing consistency among different views. In our method, the video encodings $f_{\theta}(v_1)$ and $f_{\theta}(v_2)$ are encouraged to be consistent by predicting the same latent variable $f_{\phi}(v_{ref})$.

\textbf{Effect of applying temporal pooling on the output.}
The image-based target network produces $T$ features for the input clip. Since we use the Slow pathway of~\cite{feichtenhofer2019slowfast} as our online video encoder, our online network also produces $T$ features. One natural choice is to compute the loss on each frame separately. However, we empirically observe the benefit of aggregating  feature maps of a clip before computing the loss on multiple downstream tasks.
Table~\ref{tab:temp-pool} demonstrates this when we use temporal average pooling as our aggregation function, with consistent improvement achieved across multiple target networks and downstream tasks.

\begin{table}[t]
\renewcommand{\arraystretch}{1.0}
 \small
  \floatsetup{floatrowsep=qquad, captionskip=4pt}
  \begin{floatrow}[2]
    \ttabbox%
    {\begin{tabular}{l c  c  c }
    \toprule
        Image model & $k$-NN &HMDB & UCF\\
      \midrule
       DINO-ViT & 48.0 & 59.6 & 89.6 \\ 
       DINO+SSL & 48.8 & 60.3 & 90.6\\ \midrule
       CLIP-ViT & 53.1 & 66.5 & 92.6 \\ 
       CLIP+SSL & 51.2 & 64.5 & 92.2\\ 
       \bottomrule
      \end{tabular}}
    {\caption {Effect of additional SSL loss using momentum  encoder as target network.}
      \label{tab:ssl-loss}}
     \hfill%
    \ttabbox%
    {\begin{tabular}{l c c c c }
    \toprule
       Image model & T. Pool & $k$-NN & HMDB & UCF\\

      \midrule
      DINO-ViT & \xmark & 48.0 & 59.6 & 89.6\\ 
      DINO-ViT & \cmark & 48.4 & 62.1 & 91.0\\  \midrule
      CLIP-ViT & \xmark & 53.1 & 66.5 & 92.6\\ 
      CLIP-ViT & \cmark & 54.3 & 68.6 & 94.0\\ 
      \bottomrule
      \end{tabular}}
    {\caption {Effect of applying temporal pooling on the features of online and target networks.}
      \label{tab:temp-pool}}
   
  \end{floatrow}
  \vspace{-10pt}
\end{table}

\textbf{Effect of number of temporal views.}
Here, we study the effect of the number of temporal views within a sequence. To this end, we compare three scenarios: (i) $v_1=v_{ref}$, where we use only one clip as the input to both online and target network, (ii) $v_{ref}$ and $v_1$, wherein we use $v_{ref}$ as the input to the target network and $v_{1}$ as the input to the online network, and (iii) $v_{ref}$, $v_1$, and $v_2$, wherein we use three separate clips per sequence, one acting as the input to the target network and the other two acting as the inputs to the online network (our original setting, as shown in Figure~\ref{fig:pipeline} and explained in Sec.~\ref{sec:method}). Table~\ref{tab:view} summarizes the results of these three scenarios. Results on the K400 $K$-NN classification clearly suggests having three separate views is advantageous compared to the other two cases. However, results on HMDB51 and UCF101 show on-par performance for all three scenarios. This is mainly due to the fact that spatial information is of more importance on these dataset, such clues can be learned with almost any number of views when using a strong image-based target network. To confirm this, we also add the results on SSv2 dataset, where as opposed to other datasets, temporal cues are of significant importance in discriminating between actions. The results on SSv2 suggest that adding more views, i.e., having  $v_{ref}$, $v_1$, and $v_2$, help the online video encoder to better learn temporal information.
Please note that adding more temporal views may lead to even better performance, as also investigated in~\cite{feichtenhofer2021large, corp_iccv21,brave_iccv21}, however, due to computational budget limitations, this aspect is left for future investigations. Also note, even with $v_{ref}$, $v_1$, and $v_2$, our method outperforms existing methods with more temporal views (see Table~\ref{tab:sota}).

\begin{table}[t]
\renewcommand{\arraystretch}{1.0}
 \small
  \floatsetup{floatrowsep=qquad, captionskip=4pt}
  \begin{floatrow}[2]
    \ttabbox%
    {\begin{tabular}{l c c c c }
      \toprule
      \#views & $k$-NN & HMDB & UCF & SSv2\\
      \midrule
       - & 52.7 & 68.4 & 94.5 & 53.6\\
       1 & 53.4 & 68.3 & 93.7 & 53.9 \\
       2 & 54.3 & 68.6 & 94.0 & 54.6\\
      \bottomrule
      \end{tabular}}
    {\caption {Effect of number of temporal views for the online network.}
      \label{tab:view}}

    \hfill%
    \renewcommand{\arraystretch}{1.0}
    \small
    \ttabbox%
        {\begin{tabular}{l c c c c }
      \toprule
      \#Batch & \#Epochs & $k$-NN & HMDB & UCF\\
      \midrule
        $8\times13$ & 100 & 54.3 & 68.6 &  94.0\\ 
        $32\times13$ & 100 & 55.0 & 69.1 & 94.2\\ 
        $32\times13$ & 200 & 56.2 & 71.6 & 95.0\\ 
      \bottomrule
      \end{tabular}}
    {\caption {Effect of using larger batch size and more epochs during pre-training.}
     \label{tab:bsz}}
    
  \end{floatrow}
\end{table}

\begin{table}[t]
\renewcommand{\arraystretch}{1.0}
 \small
  \floatsetup{floatrowsep=qquad, captionskip=4pt}
  \begin{floatrow}[2]
    \ttabbox%
{\begin{tabular}{l c c c}
    \toprule
      Backbone & $k$-NN & HMDB & UCF\\
      \midrule
       3DResNet-50 & 54.3 & 68.6 & 94.0\\ 
       3DResNet-101 & 55.3 & 70.1 & 94.6\\ 
    \bottomrule
      \end{tabular}}
    {\caption {Effect of using different backbone architectures for the online network.}
      \label{tab:backbone}}

    \hfill%
    \ttabbox%
    {\begin{tabular}{l c c c  c }
       \toprule
      Data & $k$-NN & K400 Lin. Eval.  & HMDB & UCF\\
      \midrule
       K400 & 54.3 & 69.1 & 68.6 & 94.0\\ 
       K600 & 55.3 & 71.4  & 69.4 & 94.5 \\ 
       \bottomrule
      \end{tabular}}
    {\caption {Effect of using a larger pre-training dataset.}
      \label{tab:k600}}
      \vspace{-5pt}
  \end{floatrow}
  \vspace{-10pt}
\end{table}

\textbf{Effect of larger batch size and number of epochs during pre-training.}
We conduct different configurations in Table~\ref{tab:bsz} in terms of batch size and number of epochs during the pre-training. 
Unsurprisingly, training for more epochs and larger batch size typically leads to an improvement in the performance of the downstream tasks. However, Table~\ref{tab:bsz} shows that our model also converges to a reasonable model even after 100 epochs (which can be considered as a small-scale experiment in video SSL). For instance, within 100 epochs our approach achieves the accuracy of 94.2\% on UCF101, outperforming the state-of-the-art~\cite{feichtenhofer2021large} by a large margin (94.2\% versus 88.6\%) with similar number of training epoch. This clearly shows the effect of using a pre-trained image-based target network in faster convergence more efficiently.

\textbf{Effect of different backbone architectures for video encoder.}
Table~\ref{tab:backbone} compares different backbone architectures for the video encoder (online branch). In particular, we compare 3DResNet-50 and 3DResNet-101 backbones. It is clear that using a deeper architecture is more effective in video SSL and improves the downstream tasks performance at the cost of more computational resources. 

\textbf{Effect of using a larger pre-training dataset.}
The results shown in Table~\ref{tab:k600} illustrate the effect of pre-training on the larger-scale K600~\cite{carreira2018short} compared to K400. As expected, this experiment indicates the benefits of using larger-scale datasets for self-supervised pre-training in all downstream tasks.

\begin{wraptable}{r}{6.7cm}
\renewcommand{\arraystretch}{1.0}
\centering
\small
\caption {Effect of video encoder initialization. Note that all baselines are trained for 200 epochs and batch size of $13\times32$.}
  \vspace{-5pt}
      \label{val2}
  \begin{tabular}{l c c  c }
  \toprule
       Method &  $k$-NN & HMDB & UCF\\
      \midrule
       $\rho$BYOL (from scratch) & 46.6 & 64.1 & 91.2\\ 
       $\rho$BYOL (inflated) & 50.1 & 65.4 & 92.2\\ 
       iBoot & 56.2 & 71.6 & 95.0\\
       \bottomrule
  \end{tabular}
  \label{tab:baseline}
\end{wraptable}

\textbf{Effect of video encoder initialization.}
Typically, video SSL methods start training the video encoders from scratch. While this is a valid solution, we also investigate the effect of starting from a better initialization for the video encoders, the result of which is shown in Table~\ref{tab:baseline}. Specifically, the second row shows the results of a model started from the inflated weights of a ResNet-50 network pre-trained on ImageNet1K with self-supervision~\cite{grill2020bootstrap}. The improvement achieved by such initialization demonstrates the effectiveness of incorporating learned image representation into the the video SSL model. While such initialization is effective, iBoot outperforms such pre-training by large margin thanks to the freedom in selecting an arbitrary target network architecture.

\begin{table}[t]
\renewcommand{\arraystretch}{1.0}
  \centering
   \small
  \caption{Comparison with state-of-the-art. (*) means averaged over 3 splits.}
  \vspace{-5pt}
  \scalebox{0.95}{
  \begin{tabular}{l c c  c c  c  c  c  }
  \toprule
  & \multicolumn{3}{c}{Pre-training}  & \multicolumn{4}{c}{Evaluation}
  \\\cmidrule(lr){2-4}\cmidrule(lr){5-8}
  \multirow{2}{*} {Method} & \multirow{2}{*}{Data} & \multirow{2}{*}{Epoch} & \multirow{2}{*}{$T\times \tau$}& {K400} & \multirow{2}{*}{UCF101} & \multirow{2}{*}{HMDB51} & \multirow{2}{*}{SSv2 (setting)}\\ & & & & lin.eval& & & 
    \\
     \midrule
     Fully Supervised &  Scratch & - & $8\times8$ & 74.7 & 42.7 & 18.4 & 48.8 (1 $v$, $8\times8$) \\
     Fully Supervised &  K400 & - & $8\times8$ & N/A & 96.6 & 76.4 & 52.8 (1 $v$, $8\times8$)\\
    \midrule
    CVRL~\cite{cvrl_cvpr21}&  K400 & 800 & $16\times2$ & 66.1 & 92.9 & 67.9 & - \\
   \rowcolor{Gray}($\rho=4$) CORP$_m$~\cite{corp_iccv21} &  K400 & 800 & $16\times2$ & 59.1 & - & - & 61.0 (30 $v$, $32\times1$)\\
    CORP$_f$~\cite{corp_iccv21} &  K400 & 800 & $16\times2$ & 66.3 & 93.5 & 68.0 & - \\
   \rowcolor{Gray} MoDist~\cite{xiao2021modist} & K400 & 600 & $8\times8$ & - & 94.0 & 67.4 & 57.4 (- $v$, $16\times8$)\\
    ($\rho=2$)BYOL~\cite{feichtenhofer2021large} & K400 & 200 & $8\times8$ & 65.8 & 92.7 & - & 54.4 (1 $v$, $8\times8$)\\
    \rowcolor{Gray} ($\rho=3$)BYOL~\cite{feichtenhofer2021large} &  K400 & 200 & $8\times8$ & 68.3 & 93.8 & - & \textbf{55.8} (1 $v$, $8\times8$) \\
    ($\rho=4$)BYOL~\cite{feichtenhofer2021large} &  K400 & 200 & $8\times8$ & 70.0 & 94.2* & 72.1* & - \\
    \rowcolor{Gray} ($\rho=3$)SimCLR~\cite{feichtenhofer2021large} & K400 & 200 & $8\times8$ & 62.0 & 87.9 & - & 52.0 (1 $v$, $8\times8$)\\
    ($\rho=3$)SwAV~\cite{feichtenhofer2021large} & K400 & 200 & $8\times8$ & 62.7 & 89.4 & - & 51.7 (1 $v$, $8\times8$)\\
    \rowcolor{Gray} ($\rho=3$)MoCo~\cite{feichtenhofer2021large} & K400 & 200 & $8\times8$ & 67.3 & 92.8 & - & 54.4 (1 $v$, $8\times8$)\\
     \midrule
     \multirow{3}{*} {iBoot} &  \multirow{3}{*}{K400} &  \multirow{3}{*}{200} &   \multirow{3}{*}{$8\times8$} &  \multirow{3}{*}{\textbf{71.8}} &
     \textbf{95.0} & 71.6  & \textbf{55.8} (1 $v$, $8\times8$) \\
     & & &  & & \textbf{94.5}*& 70.2* &  \textbf{61.5} (3 $v$, $8\times8$) \\
     & & & & & & &  \textbf{63.0} (3 $v$, $16\times8$)\\
     \bottomrule
  \end{tabular}}
   \vspace{-10pt}
  \label{tab:sota}
\end{table}

\subsection{Comparison with the state of the art}
Table~\ref{tab:sota} presents a comprehensive comparison to existing video SSL methods that, as ours, use the K400 dataset for pre-training and ResNet-50 as the video encoder backbone. This table reports the results on a number of downstream tasks including linear evaluation on K400, fine-tuning on UCF101, HMDB51 and SSv2. We emphasize that according to various experimental settings different approaches utilize, a strictly direct comparison is not feasible. To better compare to these approaches, we additionally provide the number of pre-training epochs, $T\times\tau$, and number of views ($\rho$). 

Although iBoot is trained efficiently for only 200 epochs with  $T=\tau=8$ and $\rho=2$, it outperforms existing approaches, including the ones trained for longer on more views with larger input sizes. 
As clearly shown in our results in Table~\ref{tab:sota}, thanks to its ability to capture strong semantic patterns guided by strong pre-trained image-based Vision Transformer models such as CLIP, iBoot is capable of learning strong video representations, evidenced by superior Top-1  accuracy compared to the existing methods. 
This is not only shown to be effective in spatially-heavy datasets such as K400 and UCF101, but also in the challenging datasets such as SSv2, where the temporal information is a strong cue to reason about action labels, outperforming its fully-supervised version pre-trained on K400 with a large margin. 
Apart from the methods presented in Table~\ref{tab:sota}, there exists other approaches that require drastically heavier computations. For instance, BraVe~\cite{brave_iccv21} trains four streams of video encoders for 600 epochs, each taking as input clips of either 16 or 64 frames. It also uses clips of 32 frames for all downstream tasks. While iBoot demands much less computation for pre-training and for downstream tasks, it outperforms BraVe~\cite{brave_iccv21}, e.g., achieving 95.2 (versus 94.7) on UCF101, when using 32 frames for downstream task only.

\begin{wraptable}{r}{8cm}
  \centering
  \small
  \caption{Semi-supervised classification top-1 on K400.}
  \begin{tabular}{l c c}
  \toprule
    Method & 1\% Labels & 10\% Labels\\
     \midrule
      Supervised &  4.5\% & 34.1\%\\
      SimCLR infla.~\cite{cvrl_cvpr21} & 11.8\% & 46.1\% \\
      ImageNet infla.~\cite{cvrl_cvpr21} & 16.0\% & 49.1\% \\
      CVRL~\cite{cvrl_cvpr21} & 35.1\% & 58.1\%\\ 
      CORP$_f$~\cite{corp_iccv21} & 34.8\% & 58.6\%\\
      ($\rho=2$)BYOL~\cite{feichtenhofer2021large} & 37.2\% & 60.2\%\\
      \midrule
      iBoot   &  \textbf{40.1\%}  &  \textbf{61.7\%} \\
      \bottomrule

  \end{tabular}
  \label{tab:semi-sup}
\end{wraptable}

Finally, the superior performance of iBoot is further confirmed by our experiments on the semi-supervised action recognition on K400, shown in Table~\ref{tab:semi-sup},  targeting highly challenging scenario of accessing to only 1\% and 10\% of the labeled data, as described in Section~\ref{subsec:eval}.
Note that we report the results for $\rho$BYOL~\cite{feichtenhofer2021large} in Table~\ref{tab:semi-sup} by borrowing its publicly available checkpoint \footnote{ Available at \url{https://github.com/facebookresearch/SlowFast}.} and fine-tuning it with similar setting as iBoot.

\section{Discussion}
   
\textbf{Conclusion.} 
\label{sec:discussion}
In this work, we propose iBoot, as a video representation learning pipeline, that during training encourages the video encoder to subsume the semantic content of an image-based foundation model trained on a general domain. Since, image-based models have learnt to represent images expressively, we empirically show that we can skip re-doing this for learning spatial aspects of the video. Our results on different downstream tasks have shown the benefits of relying on strong learned image representations in a self-supervised or weakly-supervised manner.

\textbf{Limitations and Societal impacts.} Despite the effectiveness of iBoot as a video representation learning framework, there are some limitations and rooms for improvement. For instance, iBoot relies on a pre-trained image foundation model, trained on an image dataset. This may introduce biases to our video SSL approach when it comes to learning spatial aspect of the videos. While this may be an important issue, further investigation on the bias introduced by the foundation model is out of scope of this work. Additionally, similar to existing video SSL approaches, iBoot is trained on Kinetics dataset comprising only trimmed videos. Evaluating the behavior of iBoot when dealing with uncurated data is yet to be seen, left for future explorations. 

On societal impact aspect, we note that iBoot gets benefits from CLIP~\cite{clip_icml21}. CLIP itself is trained on unfiltered and uncurated image-text pairs collected from the internet, resulting in learning many social biases. While a great effort has been made in the original work~\cite{clip_icml21} to provide preliminary analysis of some of these biases in the model, we hope that the future research focuses more on the shortcomings and social biases of such models.

\printbibliography

@string{ICCV = "IEEE International Conference on Computer Vision"}

@string{ECCV = "European Conference on Computer Vision"}

@string{CVPR = "IEEE Conference on Computer Vision and Pattern Recognition"}

@string{NeurIPS = "Advances on Neural Information Processing Systems"}

@string{ICLR = "International Conference on Learning Representations"}

@string{ICML = "International Conference on Machine Learning"}

@inproceedings{caron2021emerging,
    title={Emerging Properties in Self-Supervised Vision Transformers},
    author={Caron, Mathilde and Touvron, Hugo and Misra, Ishan and J\'egou, Herv\'e  and Mairal, Julien and Bojanowski, Piotr and Joulin, Armand},
    booktitle=ICCV,
    year={2021}
}

@inproceedings{grill2020bootstrap,
    title={Bootstrap your own latent: A new approach to self-supervised learning},
    author={Grill, Jean-Bastien and Strub, Florian and Altch\'{e}, Florent and Tallec, Corentin and Richemond, Pierre and Buchatskaya, Elena and Doersch, Carl and Avila Pires, Bernardo and Guo, Zhaohan and Gheshlaghi Azar, Mohammad and Piot, Bilal and Kavukcuoglu, Koray and Munos, Remi and Valko, Michal},
    booktitle=NeurIPS,
    year={2020}
}

@inproceedings{simclr_imcl20,
    title={A Simple Framework for Contrastive Learning of Visual Representations},
    author={Chen, Ting and Kornblith, Simon and Norouzi, Mohammad and Hinton, Geoffrey},
    booktitle=ICML,
    year={2020}
}

@inproceedings{simclrv2_neurips20,
    title={Big Self-Supervised Models are Strong Semi-Supervised Learners},
    author={Chen, Ting and Kornblith, Simon and Swersky, Kevin and Norouzi, Mohammad and Hinton, Geoffrey},
    booktitle=NeurIPS,
    year={2020}
}

@Article{mocov2,
    author  = {Xinlei Chen and Haoqi Fan and Ross Girshick and Kaiming He},
    title   = {Improved Baselines with Momentum Contrastive Learning},
    journal = {arXiv preprint arXiv:2003.04297},
    year    = {2020},
}

@InProceedings{HardNegMine_neurips20,
    author = {Kalantidis, Yannis and Sariyildiz, Mert Bulent and Pion, Noe and Weinzaepfel, Philippe and Larlus, Diane},
    title = {Hard Negative Mixing for Contrastive Learning},
    booktitle = NeurIPS,
    year = {2020}
}

@inproceedings{swav2020,
    title={Unsupervised Learning of Visual Features by Contrasting Cluster Assignments},
    author={Caron, Mathilde and Misra, Ishan and Mairal, Julien and Goyal, Priya and Bojanowski, Piotr and Joulin, Armand},
    booktitle=NeurIPS,
    year={2020}
}

@inproceedings{simsiam_cvpr21,
    author = {Xinlei Chen and Kaiming He},
    title  = {Exploring Simple Siamese Representation Learning},
    booktitle = CVPR,
    year      = {2021}
}

@inproceedings{clip_icml21,
    author = {Alec Radford and Jong Wook Kim and Chris Hallacy and Aditya Ramesh and Gabriel Goh and Sandhini Agarwal and Girish Sastry and Amanda Askell and Pamela Mishkin and Jack Clark and Gretchen Krueger and Ilya Sutskever},
    title     = {Learning Transferable Visual Models From Natural Language Supervision},
    booktitle = ICML,
    year      = {2021}
}

@inproceedings{cvrl_cvpr21,
    author = {Rui Qian and Tianjian Meng and Boqing Gong and Ming{-}Hsuan Yang and Huisheng Wang and Serge J. Belongie and Yin Cui},
    title     = {Spatiotemporal Contrastive Video Representation Learning},
    booktitle = CVPR,
    year      = {2021},
}

@article{vimpac_arxiv21,
  title={{VIMPAC}: Video Pre-Training via Masked Token Prediction and Contrastive Learning},
  author={Hao Tan and Jie Lei and Thomas Wolf and Mohit Bansal},
  journal={arXiv preprint arXiv:2106.11250},
  year={2021}
}

@inproceedings{beit_iclr22,
    title={{BE}iT: {BERT} Pre-Training of Image Transformers},
    author={Hangbo Bao and Li Dong and Songhao Piao and Furu Wei},
    booktitle=iclr,
    year={2022},
}

@article{zhou2021ibot,
  title={ibot: Image bert pre-training with online tokenizer},
  author={Zhou, Jinghao and Wei, Chen and Wang, Huiyu and Shen, Wei and Xie, Cihang and Yuille, Alan and Kong, Tao},
  journal={arXiv preprint arXiv:2111.07832},
  year={2021}
}

@article{he2021masked,
  title={Masked autoencoders are scalable vision learners},
  author={He, Kaiming and Chen, Xinlei and Xie, Saining and Li, Yanghao and Doll{\'a}r, Piotr and Girshick, Ross},
  journal={arXiv preprint arXiv:2111.06377},
  year={2021}
}

@article{wei2021masked,
  title={Masked Feature Prediction for Self-Supervised Visual Pre-Training},
  author={Wei, Chen and Fan, Haoqi and Xie, Saining and Wu, Chao-Yuan and Yuille, Alan and Feichtenhofer, Christoph},
  journal={arXiv preprint arXiv:2112.09133},
  year={2021}
}

@InProceedings{corp_iccv21,
    author    = {Hu, Kai and Shao, Jie and Liu, Yuan and Raj, Bhiksha and Savvides, Marios and Shen, Zhiqiang},
    title     = {Contrast and Order Representations for Video Self-Supervised Learning},
    booktitle = ICCV,
    year      = {2021},
}

@InProceedings{brave_iccv21,
    author    = {Recasens, Adri\`a and Luc, Pauline and Alayrac, Jean-Baptiste and Wang, Luyu and Strub, Florian and Tallec, Corentin and Malinowski, Mateusz and P\u{a}tr\u{a}ucean, Viorica and Altch\'e, Florent and Valko, Michal and Grill, Jean-Bastien and van den Oord, A\"aron and Zisserman, Andrew},
    title     = {Broaden Your Views for Self-Supervised Video Learning},
    booktitle = ICCV,
    year      = {2021},
}

@inproceedings{feichtenhofer2021large,
    title={A Large-Scale Study on Unsupervised Spatiotemporal Representation Learning},
    author={Feichtenhofer, Christoph and Fan, Haoqi and Xiong, Bo and Girshick, Ross and He, Kaiming},
    booktitle=CVPR,
    year={2021}
}

@misc{fan2020pyslowfast,
    author = {Haoqi Fan and Yanghao Li and Bo Xiong and Wan-Yen Lo and Christoph Feichtenhofer},
    title = {PySlowFast},
    howpublished = {\url{https://github.com/facebookresearch/slowfast}},
    year =         {2020}
}

@inproceedings{feichtenhofer2019slowfast,
    title={Slowfast networks for video recognition},
    author={Feichtenhofer, Christoph and Fan, Haoqi and Malik, Jitendra and He, Kaiming},
    booktitle=CVPR,
    year={2019}
}

@InProceedings{deit_icml21,
    title =     {Training data-efficient image transformers \& distillation through attention},
    author =    {Touvron, Hugo and Cord, Matthieu and Douze, Matthijs and Massa, Francisco and Sablayrolles, Alexandre and Jegou, Herve},
    booktitle = ICML,
    year =      {2021},
}

@inproceedings{misra2016shuffle,
    title={Shuffle and learn: unsupervised learning using temporal order verification},
    author={Misra, Ishan and Zitnick, C Lawrence and Hebert, Martial},
    booktitle=ECCV,
    year={2016},
}

@inproceedings{wei2018learning,
    title={Learning and using the arrow of time},
    author={Wei, Donglai and Lim, Joseph J and Zisserman, Andrew and Freeman, William T},
    booktitle=CVPR,
    year={2018}
}

@inproceedings{benaim2020speednet,
    title={Speednet: Learning the speediness in videos},
    author={Benaim, Sagie and Ephrat, Ariel and Lang, Oran and Mosseri, Inbar and Freeman, William T and Rubinstein, Michael and Irani, Michal and Dekel, Tali},
    booktitle=CVPR,
    year={2020}
}

@inproceedings{wang2015unsupervised,
    title={Unsupervised learning of visual representations using videos},
    author={Wang, Xiaolong and Gupta, Abhinav},
    booktitle=ICCV,
    year={2015}
}

@inproceedings{lee2017unsupervised,
    title={Unsupervised representation learning by sorting sequences},
    author={Lee, Hsin-Ying and Huang, Jia-Bin and Singh, Maneesh and Yang, Ming-Hsuan},
    booktitle=ICCV,
    year={2017}
}

@inproceedings{xu2019self,
    title={Self-supervised spatiotemporal learning via video clip order prediction},
    author={Xu, Dejing and Xiao, Jun and Zhao, Zhou and Shao, Jian and Xie, Di and Zhuang, Yueting},
    booktitle=CVPR,
    year={2019}
}

@inproceedings{wang2019learning,
    title={Learning correspondence from the cycle-consistency of time},
    author={Wang, Xiaolong and Jabri, Allan and Efros, Alexei A},
    booktitle=CVPR,
    year={2019}
}

@inproceedings{han2020memory,
    title={Memory-augmented dense predictive coding for video representation learning},
    author={Han, Tengda and Xie, Weidi and Zisserman, Andrew},
    booktitle=ECCV,
    year={2020},
}

@inproceedings{morgado2021audio,
  title={Audio-visual instance discrimination with cross-modal agreement},
  author={Morgado, Pedro and Vasconcelos, Nuno and Misra, Ishan},
  booktitle=cvpr,
  year={2021}
}

@article{alwassel2020self,
  title={Self-supervised learning by cross-modal audio-video clustering},
  author={Alwassel, Humam and Mahajan, Dhruv and Korbar, Bruno and Torresani, Lorenzo and Ghanem, Bernard and Tran, Du},
  journal=neurips,
  year={2020}
}

@inproceedings{arandjelovic2017look,
  title={Look, listen and learn},
  author={Arandjelovic, Relja and Zisserman, Andrew},
  booktitle=iccv,
  year={2017}
}

@inproceedings{hu2021contrast,
  title={Contrast and order representations for video self-supervised learning},
  author={Hu, Kai and Shao, Jie and Liu, Yuan and Raj, Bhiksha and Savvides, Marios and Shen, Zhiqiang},
  booktitle=iccv,
  year={2021}
}

@inproceedings{patrick2021space,
  title={Space-Time Crop \& Attend: Improving Cross-modal Video Representation Learning},
  author={Patrick, Mandela and Huang, Po-Yao and Misra, Ishan and Metze, Florian and Vedaldi, Andrea and Asano, Yuki M and Henriques, Jo{\~a}o F},
  booktitle=iccv,
  year={2021}
}

@inproceedings{li2021motion,
  title={Motion-Focused Contrastive Learning of Video Representations},
  author={Li, Rui and Zhang, Yiheng and Qiu, Zhaofan and Yao, Ting and Liu, Dong and Mei, Tao},
  booktitle=iccv,
  year={2021}
}

@inproceedings{behrmann2021long,
  title={Long short view feature decomposition via contrastive video representation learning},
  author={Behrmann, Nadine and Fayyaz, Mohsen and Gall, Juergen and Noroozi, Mehdi},
  booktitle=iccv,
  year={2021}
}

@inproceedings{huang2021self,
  title={Self-supervised video representation learning by context and motion decoupling},
  author={Huang, Lianghua and Liu, Yu and Wang, Bin and Pan, Pan and Xu, Yinghui and Jin, Rong},
  booktitle=cvpr,
  year={2021}
}

@article{xiao2021modist,
  title={Modist: Motion distillation for self-supervised video representation learning},
  author={Xiao, Fanyi and Tighe, Joseph and Modolo, Davide},
  journal={arXiv preprint arXiv:2106.09703},
  year={2021}
}

@inproceedings{jenni2021time,
  title={Time-equivariant contrastive video representation learning},
  author={Jenni, Simon and Jin, Hailin},
  booktitle=iccv,
  year={2021}
}

@inproceedings{lin2021self,
  title={Self-supervised video representation learning with meta-contrastive network},
  author={Lin, Yuanze and Guo, Xun and Lu, Yan},
  booktitle=iccv,
  year={2021}
}

@inproceedings{wang2021removing,
  title={Removing the background by adding the background: Towards background robust self-supervised video representation learning},
  author={Wang, Jinpeng and Gao, Yuting and Li, Ke and Lin, Yiqi and Ma, Andy J and Cheng, Hao and Peng, Pai and Huang, Feiyue and Ji, Rongrong and Sun, Xing},
  booktitle=cvpr,
  year={2021}
}

@article{korbar2018cooperative,
  title={Cooperative learning of audio and video models from self-supervised synchronization},
  author={Korbar, Bruno and Tran, Du and Torresani, Lorenzo},
  journal=neurips,
  year={2018}
}

@article{patrick2020multi,
  title={Multi-modal self-supervision from generalized data transformations},
  author={Patrick, Mandela and Asano, Yuki M and Kuznetsova, Polina and Fong, Ruth and Henriques, Joao F and Zweig, Geoffrey and Vedaldi, Andrea},
  journal={arXiv preprint arXiv:2003.04298},
  year={2020}
}

@inproceedings{piergiovanni2020evolving,
  title={Evolving losses for unsupervised video representation learning},
  author={Piergiovanni, AJ and Angelova, Anelia and Ryoo, Michael S},
  booktitle=cvpr,
  year={2020}
}

@article{ma2021contrastive,
  title={Contrastive Learning of Global and Local Video Representations},
  author={Ma, Shuang and Zeng, Zhaoyang and McDuff, Daniel and Song, Yale},
  journal=neurips,
  year={2021}
}

@article{akbari2021vatt,
  title={Vatt: Transformers for multimodal self-supervised learning from raw video, audio and text},
  author={Akbari, Hassan and Yuan, Liangzhe and Qian, Rui and Chuang, Wei-Hong and Chang, Shih-Fu and Cui, Yin and Gong, Boqing},
  journal=neurips,
  year={2021}
}

@inproceedings{patrick2021compositions,
    title={On compositions of transformations in contrastive self-supervised learning},
    author={Patrick, Mandela and Asano, Yuki M and Kuznetsova, Polina and Fong, Ruth and Henriques, Jo{\~a}o F and Zweig, Geoffrey and Vedaldi, Andrea},
    booktitle=ICCV,
    year={2021}
}

@inproceedings{sslfewshot2020,
    author = {Jong{-}Chyi Su and Subhransu Maji and Bharath Hariharan},
    title = {When Does Self-supervision Improve Few-shot Learning?},
    year = {2020},
    booktitle=ECCV,
}

@inproceedings{girdhar2019distinit,
    title={Distinit: Learning video representations without a single labeled video},
    author={Girdhar, Rohit and Tran, Du and Torresani, Lorenzo and Ramanan, Deva},
    booktitle=ICCV,
    year={2019}
}

@inproceedings{wu2018unsupervised,
  title={Unsupervised feature learning via non-parametric instance discrimination},
  author={Wu, Zhirong and Xiong, Yuanjun and Yu, Stella X and Lin, Dahua},
  booktitle=cvpr,
  year={2018}
}

@article{kay2017kinetics,
  title={The kinetics human action video dataset},
  author={Kay, Will and Carreira, Joao and Simonyan, Karen and Zhang, Brian and Hillier, Chloe and Vijayanarasimhan, Sudheendra and Viola, Fabio and Green, Tim and Back, Trevor and Natsev, Paul and others},
  journal={arXiv preprint arXiv:1705.06950},
  year={2017}
}

@article{carreira2018short,
  title={A short note about kinetics-600},
  author={Carreira, Joao and Noland, Eric and Banki-Horvath, Andras and Hillier, Chloe and Zisserman, Andrew},
  journal={arXiv preprint arXiv:1808.01340},
  year={2018}
}

@article{devlin2018bert,
  title={Bert: Pre-training of deep bidirectional transformers for language understanding},
  author={Devlin, Jacob and Chang, Ming-Wei and Lee, Kenton and Toutanova, Kristina},
  journal={arXiv preprint arXiv:1810.04805},
  year={2018}
}

@inproceedings{ermolov2021whitening,
  title={Whitening for self-supervised representation learning},
  author={Ermolov, Aleksandr and Siarohin, Aliaksandr and Sangineto, Enver and Sebe, Nicu},
  booktitle={International Conference on Machine Learning},
  pages={3015--3024},
  year={2021},
  organization={PMLR}
}

@inproceedings{zbontar2021barlow,
  title={Barlow twins: Self-supervised learning via redundancy reduction},
  author={Zbontar, Jure and Jing, Li and Misra, Ishan and LeCun, Yann and Deny, St{\'e}phane},
  booktitle={International Conference on Machine Learning},
  pages={12310--12320},
  year={2021},
  organization={PMLR}
}

@inproceedings{larsson2016learning,
  title={Learning representations for automatic colorization},
  author={Larsson, Gustav and Maire, Michael and Shakhnarovich, Gregory},
  booktitle={European conference on computer vision},
  pages={577--593},
  year={2016},
  organization={Springer}
}

@inproceedings{zhang2016colorful,
  title={Colorful image colorization},
  author={Zhang, Richard and Isola, Phillip and Efros, Alexei A},
  booktitle={European conference on computer vision},
  pages={649--666},
  year={2016},
  organization={Springer}
}

@inproceedings{pathak2016context,
  title={Context encoders: Feature learning by inpainting},
  author={Pathak, Deepak and Krahenbuhl, Philipp and Donahue, Jeff and Darrell, Trevor and Efros, Alexei A},
  booktitle={Proceedings of the IEEE conference on computer vision and pattern recognition},
  pages={2536--2544},
  year={2016}
}

@inproceedings{goyal2017something,
  title={The" something something" video database for learning and evaluating visual common sense},
  author={Goyal, Raghav and Ebrahimi Kahou, Samira and Michalski, Vincent and Materzynska, Joanna and Westphal, Susanne and Kim, Heuna and Haenel, Valentin and Fruend, Ingo and Yianilos, Peter and Mueller-Freitag, Moritz and others},
  booktitle={Proceedings of the IEEE international conference on computer vision},
  pages={5842--5850},
  year={2017}
}

@article{soomro2012ucf101,
  title={UCF101: A dataset of 101 human actions classes from videos in the wild},
  author={Soomro, Khurram and Zamir, Amir Roshan and Shah, Mubarak},
  journal={arXiv preprint arXiv:1212.0402},
  year={2012}
}

@InProceedings{Kuehne11,
   author= "Kuehne, H. and Jhuang, H. and Garrote, E. and Poggio, T. and Serre, T.",
   title = "{HMDB}: a large video database for human motion recognition",
   booktitle = "Proceedings of the International Conference on Computer Vision (ICCV)",
   year = "2011",
}

@article{van2017neural,
  title={Neural discrete representation learning},
  author={Van Den Oord, Aaron and Vinyals, Oriol and others},
  journal={Advances in neural information processing systems},
  volume={30},
  year={2017}
}

@article{van2018representation,
  title={Representation learning with contrastive predictive coding},
  author={Van den Oord, Aaron and Li, Yazhe and Vinyals, Oriol},
  journal={arXiv e-prints},
  pages={arXiv--1807},
  year={2018}
}

@inproceedings{gidaris2019boosting,
  title={Boosting few-shot visual learning with self-supervision},
  author={Gidaris, Spyros and Bursuc, Andrei and Komodakis, Nikos and P{\'e}rez, Patrick and Cord, Matthieu},
  booktitle={Proceedings of the IEEE/CVF International Conference on Computer Vision},
  pages={8059--8068},
  year={2019}
}

@inproceedings{zhai2019s4l,
  title={S4l: Self-supervised semi-supervised learning},
  author={Zhai, Xiaohua and Oliver, Avital and Kolesnikov, Alexander and Beyer, Lucas},
  booktitle={Proceedings of the IEEE/CVF International Conference on Computer Vision},
  pages={1476--1485},
  year={2019}
}

@inproceedings{chen2021pareto,
  title={Pareto self-supervised training for few-shot learning},
  author={Chen, Zhengyu and Ge, Jixie and Zhan, Heshen and Huang, Siteng and Wang, Donglin},
  booktitle={Proceedings of the IEEE/CVF Conference on Computer Vision and Pattern Recognition},
  pages={13663--13672},
  year={2021}
}

@inproceedings{wu2020self,
  title={Self-supervised domain-aware generative network for generalized zero-shot learning},
  author={Wu, Jiamin and Zhang, Tianzhu and Zha, Zheng-Jun and Luo, Jiebo and Zhang, Yongdong and Wu, Feng},
  booktitle={Proceedings of the IEEE/CVF Conference on Computer Vision and Pattern Recognition},
  pages={12767--12776},
  year={2020}
}

@article{you2017large,
  title={Large batch training of convolutional networks},
  author={You, Yang and Gitman, Igor and Ginsburg, Boris},
  journal={arXiv preprint arXiv:1708.03888},
  year={2017}
}

@article{goyal2017accurate,
  title={Accurate, large minibatch sgd: Training imagenet in 1 hour},
  author={Goyal, Priya and Doll{\'a}r, Piotr and Girshick, Ross and Noordhuis, Pieter and Wesolowski, Lukasz and Kyrola, Aapo and Tulloch, Andrew and Jia, Yangqing and He, Kaiming},
  journal={arXiv preprint arXiv:1706.02677},
  year={2017}
}

@article{loshchilov2016sgdr,
  title={Sgdr: Stochastic gradient descent with warm restarts},
  author={Loshchilov, Ilya and Hutter, Frank},
  journal={arXiv preprint arXiv:1608.03983},
  year={2016}
}

\clearpage

\section{Appendix}
    In Algorithm~\ref{alg:cap}, we provide the pseudo-code of iBoot training loop. 
Following our discussion in Section 3 of the main paper, given an image-based foundation model, we utilize different views, $v_{ref}$, $v_{1}$, and  $v_{2}$, randomly sampled and augmented from a sequence. While $v_{ref}$ acts as the input to the target network, we pass $v_{1}$ and $v_{2}$ to the online network and we simply encourage the features of $v_{1}$ and $v_{2}$ produced by the online network to match that of the target one by minimizing the cosine distance loss.

\begin{algorithm}[H]
\caption{iBoot Training Loop}\label{alg:cap}
\begin{flushleft}
\hspace*{\algorithmicindent} {\textbf{Input}:}
\end{flushleft} 
\hspace*{\algorithmicindent} $f_{\theta}$  \hspace*{\algorithmicindent}\Comment{online network (video encoder)}\\
\hspace*{\algorithmicindent} $f_{\phi}$ 
\hspace*{\algorithmicindent}\Comment{target network (image-based pre-trained model)}\\
\hspace*{\algorithmicindent} $W$ 
\hspace*{\algorithmicindent}\Comment{linear transformation mapping}\\
\begin{algorithmic}
\For{\texttt{$v$ in loader}}  \Comment{load a minibatch $v$ with $n$ samples, each with $T$ frames}
        \State \texttt{$v_{ref}, v_1, v_2 = \mathcal{T}(v), \mathcal{T}(v), \mathcal{T}(v) $} 
        \Comment{$\mathcal{T}$ is a random set of spatio-temporal augmentations}
        \State \texttt{$v_{1-2}$ = Concat([$v_1$,$v_2$])}
        \State \texttt{$k = W(f_{\theta}(v_{1-2}))$}
        \State \texttt{$v_{ref}$ = Reshape$(v_{ref})$} 
        \hspace*{\algorithmicindent}\Comment{$n,T \rightarrow n \times T$}
        \State \texttt{$q = f_{\phi}(v_{ref})$}
        \hspace*{\algorithmicindent}\Comment{output \texttt{[CLS]} token if $f_\phi$ is ViT model}
        \State \texttt{$q$ = Reshape$(q)$} 
        \hspace*{\algorithmicindent}\Comment{$n \times T \rightarrow n,T$}
        \State \texttt{$q$ = Concat([$q$,$q$])}
        \State \texttt{loss = $2-2.\frac{\langle q \; , \; k \rangle }{\| q \|_2. \|k\|_2}$}
        \State \texttt{loss.backward()}
        \State \texttt{update($f_{\theta}$)}
\EndFor
      
\end{algorithmic}
\end{algorithm}

\end{document}